\documentclass[letterpaper, 10 pt, conference]{ieeeconf}

\usepackage{graphicx}
\usepackage{amsmath}
\usepackage{amssymb}
\usepackage{booktabs}
\usepackage{multirow}
\usepackage{subfigure}

\usepackage[dvipsnames,table,xcdraw]{xcolor}
\usepackage[ruled,vlined,linesnumbered, boxed]{algorithm2e}

\usepackage{yhmath}
\usepackage{cite}

\makeatletter
\let\NAT@parse\undefined
\makeatother
\usepackage[backref, colorlinks, linkcolor=red, anchorcolor=blue, citecolor=blue]{hyperref}

\usepackage{listings} % File listings, with syntax highlighting
\usepackage{bm}

\usepackage[framemethod=tikz]{mdframed} % Allows defining custom boxed/framed environments
\mdfdefinestyle{file}{
	innertopmargin=0.8\baselineskip,
	innerbottommargin=0.1\baselineskip,
	topline=false, bottomline=false,
	leftline=false, rightline=false,
	leftmargin=0.0cm,
	rightmargin=0cm,
	singleextra={%
		\draw[fill=black!10!white](P)++(0,-1.2em)rectangle(P-|O);
		\node[anchor=north west]
		at(P-|O){\rmfamily\mdfilename};
		\def\l{0.5em}
		\draw(O-|P)++(-\l,0)--++(\l,\l)--(P)--(P-|O)--(O)--cycle;
		\draw(O-|P)++(-\l,0)--++(0,\l)--++(\l,0);
	},
	nobreak,
}

\vspace{0.8cm}  %调整图片与上文的垂直距离

\IEEEoverridecommandlockouts                              % This command is only needed if 
\title{\LARGE \bf
Star-Convex Constrained Optimization for Visibility Planning \\with Application to Aerial Inspection
}
\author{Tianyu Liu$^{1, 2, 3}$, Qianhao Wang$^{1, 2}$, Xingguang Zhong$^{1,2} $,\\  Zhepei Wang$^{1, 2} $,  Chao Xu$^{1, 2} $, Fu Zhang$^{3} $ and Fei Gao$^{1, 2} $
	\thanks{1 State Key Laboratory of Industrial Control Technology, Institute of Cyber-Systems and Control, Zhejiang University, Hangzhou, 310027, China.} 
	\thanks{2 Huzhou Institute of Zhejiang University, Huzhou, 313000, China.}
	\thanks{3 Department of Mechanical Engineering, The University of Hong Kong.}
	\thanks{This work was supported by the National Natural Science Foundation of China under Grants 62003299.}
	\thanks{
		Corresponding author: Fei Gao, {\tt\small fgaoaa@zju.edu.cn}}
}

\begin{document}

\maketitle
\thispagestyle{empty}
\pagestyle{empty}

%%%%%%%%%%%%%%%%%%%%%%%%%%%%%%%%%%%%%%%%%%%%%%%%%%%%%%%%%%%%%%%%%%%%%%%%%%%%%%%%
\begin{abstract}
The visible capability is critical in many robot applications, such as inspection and surveillance, etc. Without the assurance of the visibility to targets, some tasks end up not being complete or even failing.
 In this paper, we propose a visibility guaranteed planner by \textit{star-convex constrained optimization}. The visible space is modeled as star convex polytope (SCP) by nature and is generated by finding the visible points directly on point cloud.
By exploiting the properties of the SCP, 
the visibility constraint is formulated for trajectory optimization. The trajectory is confined in the safe and visible flight corridor which consists of convex polytopes and SCPs. We further make a relaxation to the visibility constraints and transform the constrained trajectory optimization problem into an unconstrained one that can be reliably and efficiently solved.
To validate the capability of the proposed planner, 
we present the practical application in site inspection.
The experimental results show that the method is efficient, scalable, and visibility guaranteed, presenting the prospect of application to various other applications in the future.

%However, unlike the criterion of collision-free, dynamic feasibility and energy efficiency， the visible capibility is seldom addressed in the trajectory planning. Although there are some works take the visibility into account in trajectory
%planning, the lack of the visible capibility guarantee will degrade the planning performance.

%Drones have greatly broadened the range of information acquisition in our daily lives due to their superior mobility and agility.
\end{abstract}
%Drones have played more and more important role in many scenarios, such as inspection and surveillance.
\section{Introduction}
In many applications, such as inspection and surveillance,
enabling a drone to adjust its motion to keep interesting objects visible has high priority. Many tasks even put forward a strict demand on visibility. 
For instance, in substation inspection and factory security patrolling, specific positions must be repeatedly observed one by one in large-scale scenes. The tasks are regarded as unsuccessful or failed if any prescribed position is left unobserved. Therefore, visibility is a key constraint while designing a trajectory planner for these applications.

% an aerial tracking fail easily once the target is lost, and a site inspection is regarded as unsuccessful if any interested points left unobserved. 

%Due to their superior mobility and agility, drones greatly broaden the range of information acquisition in our daily lives.
%In many scenarios, such as substation inspection and bridge crack detection, human workers must repeatedly observe specific positions one by one in large-scale scenes, which is often time-consuming, laborious, and dangerous.  And recently,  UAVs (unmanned aerial vehicle) showed their superior mobility and agility in more and more real applications[][], so it is a natural idea to use autonomous UAVs to accomplish these tasks instead of humans.  In order to achieve this goal, the following issues need to be considered: First, During the flights, the UAVs need to stay in the visible areas of each observation position for enough time. Second, The UAVs must fly safely in large-scale and complex environments.  And Finally, the UAV's trajectory should save time and energy as much as possible to facilitate multiple missions.

% especially where complex obstacles often occlude the target perception,

Despite the significance of visibility,  most works \cite{bonatti2018autonomous, wang2021visibility, jeon2020integrated}  in the trajectory planning literature are not able to have a guarantee on it. 
Typically, they treat the visibility as an utility and optimize a handcrafted visibility cost along with other terms such as smoothness. 
However, such a formulation may trade off visibility for a smoother motion, which results in soft visibility constraints. 
Another work\cite{zhang2018perception} deterministically generates motion primitives and selects the best one among them. 
Although this method ensures visibility in a resolution complete manner, it inherently suffers from the discretization error and the curse of dimension, which cannot generate an optimal trajectory with pleasing maneuverability.
%\begin{figure}
%	\centering
%	\includegraphics[width=0.2\linewidth]{figures/front_test.png}
%	\caption{A composite image of the real world experiment. The cylindrical objects are to be inspect, which are marked by yellow bounding box.}
%	\label{front_test-}
%\end{figure}
\begin{figure}
	\centering
	\includegraphics[width=1.0\linewidth]{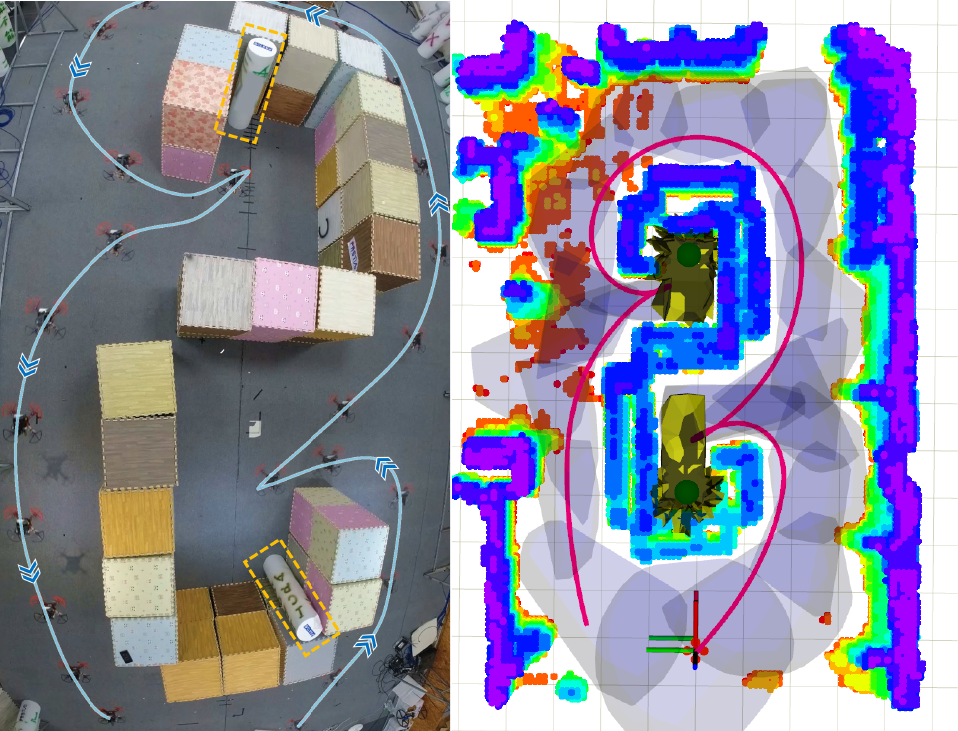}
	\caption{\textbf{Left}: a composite image of the real-world experiment in the view of fisheye cameras. The cylindrical objects are to be inspected, which are marked by yellow bounding boxes. \textbf{Right}: Illustration of the proposed visibility planner. The SCPs (yellow meshes) are constructed at the inspect spot (green dot). They form the SVC together with the convex polytopes (blue meshes) to ensure the visibility and safety of the trajectory (red line). }
	\label{front_test}
\end{figure}

To bridge the above gap, this paper proposes a planner that can efficiently generates a trajectory with visibility assurance. 
To generalize to various applications, we define the \textit{task representative points} (TRPs), which refer to the sites for inspection, the frontiers for the exploration, the places for surveillance, etc. Central to our approach is the visible space representation w.r.t. the TRPs and the corresponding constraint formulation. As we know, 
the line-of-sights from a point naturally form a star-shaped region. Based on this idea, we model the visible space as the star convex polytope (SCP), a compact and analytic representation. By utilizing the property of the constructed SCP, we formulate visibility constraint to facilate \textit{star-convex constrained optimization}.

%\begin{figure}[htbp]
%	\centering
%	\begin{minipage}[t]{0.48\textwidth}
%		\centering
%		\includegraphics[width=1.0\linewidth]{test1.jpg}
%		\caption{World Map}
%	\end{minipage}
%	\begin{minipage}[t]{0.48\textwidth}
%		\centering
%		\includegraphics[width=1.0\linewidth]{test2.jpg}
%		\caption{Concrete and Constructions}
%	\end{minipage}
%\end{figure}

%\begin{figure}
%	\caption{Differences Before and Aftrt Applying PAA} 
%	\label{front_tst}
%	\centering
%	\subfigure[123]{
%		\begin{minipage}{0.5\textwidth} %[b]%{0.2\textwidth} 
%			%{12cm or 0.2/textwidth} 控制图片大小，可以=textwidth
%			% 插入子图片
%			\includegraphics[width=\textwidth]{figures/real1.png} \\
%			%\includegraphics[width=\textwidth]{fig/beforePAA.eps} 
%			% 可以在一个minipage里写多张图片，它们共用一个小标题\subfigure['sub_title']
%		\end{minipage}
%	}
%	% 百度里有个方法说这里要空格，但我不空格也是上下排版的，疑惑
%	\subfigure[Sequence After PAA]{
%		\begin{minipage}{0.5\textwidth}%[b]%{0.2\textwidth}
%			\includegraphics[width=\textwidth]{figures/env_1_p_s.png} \\
%			
%		\end{minipage}
%	}
%\end{figure}

%suffers from a low computing efficiency. 

% which called \textit{visibility-aware trajectory planning}

In summary, the proposed planner optimizes trajectory in a \textit{safe and visible corridor} (SVC) which encodes visibility and safety requirements. The SCPs, accounting for the visibility constraints, make parts of it. The SVC is constructed by connecting all the SCPs by convex polytopes. The whole process runs in three steps. Firstly, the global optimal tour (i.e., the visiting sequence of the SCPs) is found and refined on SCPs.  Secondly, the kinodynamic A* path searching is conducted to find a collision-free path. Finally, the corridor is constructed incrementally by connecting all the SCPs with sequences of overlapping convex polytopes utilizing the searched trajectory. With the constructed SVC, we follow the work of \cite{wang2021geometrically} to optimize the trajectory spatially and temporally. The visibility constraint is further relaxed to convert the optimization problem into an unconstrained one that can be solved reliably and efficiently. To validate the planner, we apply it in the task of aerial inspection. Benchmark results show that our method is light-weighted, efficient, and scalable.
%\begin{figure}
%	\centering
%	\includegraphics[width=0.8\linewidth]{figures/principle1.pdf}
%	\caption{A similar picture is needed here.}
%	\label{front}
%\end{figure}
To conclude, the contributions of this paper are as follows:
\begin{enumerate}
	\item Introduce a new visible space representation the star-convex polytope (SCP) and propose to formulate the visibility constraint for \textit{star-convex constrained optimization}.
	\item Propose a visibility guaranteed planning framework, while retains the safety, feasibility, and energy efficiency of trajectory.
%	 with path seaching and SVC constructoin directly on point cloud, to optimize for  collison free, dynamically feasible, visibility capbale, time and energy efficient trajectory.
	\item  Validate the proposed method by implementing simulation and real-world experiments in aerial inspection.
\end{enumerate}

\section{Related Work}
%Trajectory generation for visibility, which takes the occlusion effect against obstacles into consideration, is of vital importance in various applications like inspection and tracking.
\subsection{Trajectory Planning with Visibility}\
Many works \cite{jeon2019online, jeon2020integrated} in trajectory planning design the visibility metric utilizing the minimum value of the Euclidean Signed Distance Field (ESDF) on the line between the TRPs and the robot.
Since the metric is not differentiable, they use a sampling-based method to handle the metric in trajectory generation, which is time-consuming.
Wang et al. \cite{wang2021visibility} propose a differentiable metric and yet it lacks a strong guarantee on visibility because the trajectory optimization trades off many costs.
Instead of explicitly optimize visibility, Zhou et al. \cite{zhou2021raptor} present a perception-aware strategy.
Nevertheless, the task-specific method can hardly be extended to other scenarios. Zhou et al. \cite{zhou2021fuel} propose an efficient exploration framework that naturally adapts to inspection tasks, whereas they only consider visibility in the sampling-based front-end. In this paper, we efficiently extract the visible space by SCP to facilitate trajectory planning.
%Trajectory optimization with visibility, which take the occlusion effect against obstacles into consideration, is  necessary in various applicaiton like inspection and tracking. As aforementioned, most works \cite{bonatti2018autonomous, wang2021visibility, jeon2020integrated}  design the visibility cost and minimize it along with other costs like smoothness, which lack a strong guarantee on the visibility.
%In addition, they commonly define the  occlusion cost as the integral over a 2D manifold which is built by connecting the \textit{task representative points} and drone's positions in time, which is indeed a natural idea. However, this way tackle the visibility problem from the viewpoint of robot and makes the trajectory optimization problem too much redundancy. Similar to out pipline, Jeon et al. \cite{jeon2020integrated} relief the optimization by search a graph with safety and occlusion consideration and obtain a corridor for trajectory generation. However, the corridor is still ambiguous for the safety and visibility constraints. Different from all of the existing methods, we explicitly extract the visibility spaces on the \textit{task representative points} to render a clean optimiztion problem.

\subsection{Trajectory Planning For Quadrotor}
%The point cloud map retains the most fidelity of the sensor measurements, and saves a lot of computational overhead without maintaining extra data strucure such as ESDF. To generate the trajectory on it, some works \cite{viswanathan2020efficient, zhang2019maximum} discretize the space of
%possible trajectories into motion primitives and query the collision status from the map. However, such kind of method leads to low-quality trajectories. Others abstract the environment by the safe corridor to facilitate the convex formulate of the trajectory generation problem. Liu et al. \cite{liu2017planning} propose a method to generate the corridor by convex decomposition but it need the extra grid map to find the leading path. Ji et al. \cite{ji2020mapless} propose an efficient path searching method with k-d tree, while the corridor generated is of ball shape, which impose much conservativeness on the trajectory. 
Trajectory planning for quadrotors can be categorized into the soft-constrained  and hard-constrained approaches. The former formulates the trajectory generation as NLP to trade off several objectives, but they usually suffer from the issue of local minima\cite{oleynikova2016continuous}. By exploiting the properties of B-splines, Zhou et al.\cite{zhou2019robust} propose a method but the construction of ESDF is time-consuming, especially for the large-scale trajectory planning. While  an ESDF-free planner is proposed \cite{zhou2020ego}, but the trajectory generated highly rely on and limit to the collision-
free guiding path. The
hard-constrained method is pioneered by \cite{mellinger2011minimum} which form the problem as quadratic programming (QP) with trajectory represented as piecewise polynomials. The safety can be ensured by extracting convex safe regions\cite{liu2017planning}. To obtain more reasonable time allocation, alternating minimization \cite{wang2020alternating} and mixed integer-based\cite{tordesillas2019faster} based approach are proposed. Recently, Wang et al. \cite{wang2021geometrically} proposed a spatial
and temporal optimization-based framework, which efficiently
handles a wide variant of constraints. We follow the work \cite{wang2021geometrically} for trajectory optimization in this paper. 

\section{Problem Statement}
Consider a list of TRPs in 3D space  $ \mathbb{C} = \{ c_i \in \mathbb{R}^3 |1 \le i \le N \} $. The robot starting from the position $ p_s \in \mathbb{R}^3 $ is expected to inspect all of the points in $ \mathbb{C} $ one by one and finally rest at the desired postion $ p_f \in \mathbb{R}^3 $. Commonly, the duration of inspection for each point is required to be last for at least a specific time $ \mathbb{T} = \{\tau_i \in \mathbb{R} | 1 \le i \le N\} $. For an abitrary point in $ \mathbb{R}^3 $, $ c_i $ is supposed to be visible to it if the line segment from the point to $ c_i $ is collision free. Denote $ \mathbb{S}_i \subseteq \mathbb{R}^3 $ form the space where the point $ v_i $ is visible. Since the occulusion effect against obstacles i.e. the visibility is the focus of this paper, we make the following assumptions:
\begin{enumerate}
\item  The sensor mounted on the robot has omnidirectional coverage,  which is one kind of set up of UAVs and has certain research works \cite{gao2020autonomous}.
\item  The visibility condition is satisfied only when the whole body of the robot is in the ball-shaped sensible regions around the points $ \mathbb{C} $.
\end{enumerate}

\section{Visible Space Representation}
%For the TRPs $\mathbb{C}$ to be seen, we need to obtain the space $ \mathbb{S} $ where the sites are visible to the robot. However,
Normally,
 constructing a star-shape visible region on the point cloud map is non-trivial. Collision checking of the rays starting from the sites $ c_i $ to the space needs either frequent kd-tree queries or discretization of the space. Apparently, these kinds of straightforward solutions are arduous and time-consuming. Inspired by \cite{zhong2020generating, katz2007direct, katz2015visibility}, we introduce a new method to construct visibility space represented by star convex polytope, with the emphasis on compactness and efficiency.

%by finding the direct visible point sets.

%\cite{zhong2020generating,katz2007direct}

\subsection{Star Convex Polytope Construction}
In this paper, the obstacles are represented by point cloud map $ \mathcal{M}_g $ which is organized in k-d tree structure. Our method to construct star convex polytope on $ \mathcal{M}_g $ is composed of four steps: 1) point retrieval and augment, 2) point transformation, 3) convex hull construction, 4) inversion. The main idea of the method is to find the visible points by point transformation.

In order to construct the star-shaped region within a sphere boundary with radius $ R $, we retrieve the local point cloud $ \mathcal{M}_v $ surround the point $ c_i $ by the range query on $ \mathcal{M}_g $. In addition, augmented points, which are evenly sampled on the sphere boundary, are added to better facilitate the construction.
%Illustaration of star convex polytope construction in 2D. Point transformation: the points(green) inside the sphere is transfered to the points outside of the sphere(blue) by one to one correspondence. Convex hull construction: the convex hull is constructed in the nonlinear space outside of the sphere; the blue dotted curve is the image of it in the original space. 
%\vspace{1cm}
%\smallskip
\begin{figure}[htp]
	\vspace{0.2cm}
	\centering
	\includegraphics[scale=0.80]{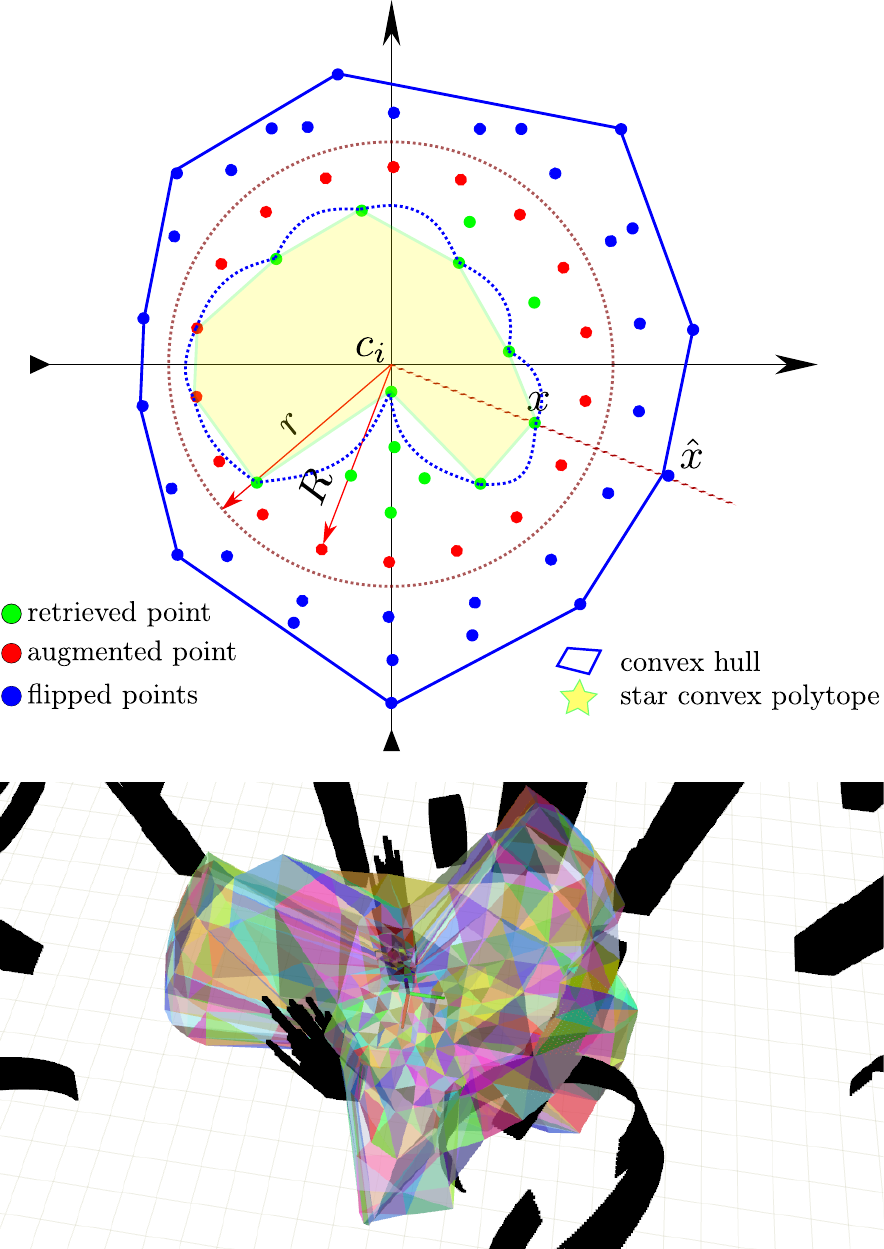}
%	\hspace{1in}
	\caption{\textbf{Top}: Illustration of SCP construction in 2D. The blue dotted curve is the inversion of the convex hull.  \textbf{Bottom}: SCP visualization in 3D as colorful mesh. It is generated on point cloud with $ R=6m $.}
	\label{scp}
\end{figure}

%\begin{figure}
%	\centering
%	\includegraphics[scale=0.4]{figures/3d_scp_t.png}
%	\caption{SCP visualization in 3D as colorful mesh. It is generated on point cloud with a confined ball region of 6m.}
%	\label{scp-3d}
%\end{figure}

%\begin{figure}[h]
%	\centering
%	\includegraphics[width=0.5\linewidth]{figures/scp-con.png}
%	\caption{A similar picture is needed here.}
%	\label{scp}
%\end{figure}

%we can perform the operation termed \textit{Hidden Point Removal}. As illustrated in , the operation consists of two steps: \textit{point transfromation} and \textit{convex hull construction} . The transfromation transfer , which is in onlinear space ouside of the sphere. 

With the point set $ \mathcal{M}_v $ and center $ c_i $, we perform point transformation that flip all the points to outside of the sphere boundary. As shown in the \hyperref[scp]{Figure \ref{scp}}, the point $ x $ is transfer to $ \hat{x} $ along the ray $ \overrightarrow{c_i x} $. The corresponding function is suppose to be monotonically decreasing. Here, we simply use the ball flipping function with ball radius $ r $: 
\begin{equation}\label{ball-flipping}
\hat{x} = F(x) = x - c_i + 2(r - \| x - c_i \| ) \frac{x - c_i}{\|x - c_i\|}.
\end{equation}

Then, we calculate the convex hull of the flipped points by the efficient convex hull algorithm \cite{barber1996quickhull}. Inherently, points that lie on the convex hull are the images of the visible points. Similarly, the convex hull is the image of the underlying star-shaped boundary of visible space. Thus we can obtain the SCP by applying the inversion of (\ref{ball-flipping}) on the convex hull and denote it as $ \mathcal{S}_i $. %The 3d illustration of the generation is show in \hyperref[scp-3d]{Figure \ref{scp-3d}}.
%The whole star convex polytope construction method is summarize in \hyperref[scp-construction]{ Algorithm 1}.
%Morever, the SCP has several notable properties:
Morever, the point that can be mapped outside the convex hull is bound to be visible by $ c_i $. The Point-In-SCP check can be performed by checking whether the flipping of the point is outside of the convex hull.

\subsection{Star-Convex Constrained Optimization}
%By utilizing the companpion 
%In order to incorporate the star convex polytope in the optimization problem \ref{formulation}, the star convex constraint can be generally formulated as:
%The visibility planning entails the study that constraint a point $ x $ in the SCP. 

The visibility planning entails the study of the following optimization problem: 
\begin{equation}
\begin{aligned}
\min_{x} \quad & \mathcal{J}(x), \quad
\mathrm{s.t.} &  x \in \mathcal{S}_i,
\end{aligned}
\label{star_opt}
\end{equation}
where $ \mathcal{J}(x) $ is the user-defined cost function.
Suppose the SCP is closed by $ K $ faces in $ \mathbb{R}^3 $.
Instead of considering it directly, 
the flipped convex polytope $ \mathcal{P}_i $ is utilized. By the  $ \mathcal{H}- $representation of convex polytope, it can be defined as
\begin{equation}
\mathcal{P}_i = \{ x \in \mathbb{R}^3 |  \mathbf{A}x \preceq \mathbf{b} \},
\end{equation}
where the matrix $ \mathbf{A} = [n_1^T, \cdots, n_K^T]^T \in \mathbb{R}^{K\times3} $ is build by the outer normal vectors of each face $ n_i, i = 1, \cdots, K $ and $ \mathbf{b} = [n_1^Ta_1, \cdots, n_K^Ta_K] \in \mathbb{R}^K $ is formed by arbitrary points $ a_i $ on each faces.
 By the property of SCP, the visibility constraint is equivalent to the insurance that the flipped point $ \hat{x} $ is outside of $ \mathcal{P}_i $, which is expressed as
\begin{equation}
\Xi \big(\hat{x}) > d_{\min},
\label{star_con_p}
\end{equation}
where $ d_{\min} $ is the user-defined safe margin and $\Xi(\cdot)$ is the signed distance function on $ \mathcal{P}_i $. The signed distance equals to zero on the surface of the convex hull. The inside and the outside of it correspond to the negative and positive Euclidean distances respectively.
 To be more specific, the signed distance is defined as
\begin{equation}
\Xi(\hat{x}) = \max_i \Big\{ d_i =  n_i^T(\hat{x} - a_i) \Big| i = 1, 2, \cdots K \Big\}.
\label{max}
\end{equation} 
 However, the maximum function introduces the non-smooth gradient and keep it away from the efficient solution of the optimization with sophisticated solvers. 
 To resolve this issue, we turn to enforce the point visibility constraint via smooth approximation of the maximum function.
% As a matter of fact, the point visibility constraint can be enfored via smooth approximation of the
%maximum function.
 Inspired by \cite{lutz2021efficient}, we employ the log-sum-exp function to make the approximation.  Thus, (\ref{max}) can be written as 
% Denote $ d_i = n_i^T(\hat{x} - a_i) $ for all $ i = 1, \cdots, K $. The (\ref{max}) can be written as 
\begin{equation}
\Xi(\hat{x})=
LSE\big(d_1, \cdots, d_K\big) = \frac{1}{\alpha}log \big(e^{\alpha d_1}+ \cdots + e^{\alpha d_K} \big),
\end{equation}
where the $ \alpha \in \mathbb{R}^{+} $ is an adjustable variable that can 
control the quality of the approximation, with $ LSE \big(d_1, d_2, \cdots, d_K \big) \rightarrow \max\big(d_1, d_2, \cdots, d_K\big) $ for $ \alpha \rightarrow +\infty $. 
%Thus we may reformulate (\ref{star_con_p}) as 
%\begin{equation}
%\frac{1}{\alpha}log\big(e^{\alpha (n_i^T(p - a_i))}+ \cdots + e^{\alpha (n_i^T(p - a_i))}\big) > d_{min}
%\label{star_con_re}
%\end{equation}
Furthermore, we relax the original optimization problem (\ref{star_opt}) by constraint violation to convert it into an unconstrained problem:
\begin{equation}
\begin{aligned}
\min_{x} \quad  \mathcal{J}(x) +  \mathcal{V}(\widehat{LSE}), \\ 
\end{aligned}
\label{star_opt_uncon}
\end{equation}
where 
\begin{equation}
\mathcal{V}(\widehat{LSE}) = \lambda \max \big\{\widehat{LSE}, 0 \big\}^3.
\label{scp_v}
\end{equation}
The $ \lambda \in \mathbb{R}^+ $ is the an extremely large penalty weight and the $ \widehat{LSE} $ stand for
\begin{equation}
\begin{aligned}
\widehat{LSE} \big(\hat{x} | \mathcal{S}_i \big) =  d_{\min} - \frac{1}{\alpha} log \big( \sum_{i = 1}^{K} e^{\alpha d_i } \big).
\end{aligned}
\label{lsehat}
\end{equation}
Apparently, the violation term (\ref{scp_v}) preserves the $ C^2 $ condition, making the second order gradient attainable. Given the visible space $ \mathcal{S}_i $, 
 we can derive the gradient of $ \mathcal{V} $ w.r.t. $ x $ from (\ref{ball-flipping}), (\ref{scp_v}) and (\ref{lsehat}) and denote it by $ g_{scp} $. The gradient is zero when $ \widehat{LSE} \le 0  $, and for $ \widehat{LSE} > 0 $, the gradient is given as
\begin{equation}
g_{scp}=
\frac{\partial \mathcal{V}}{\partial x} = 
6 \lambda {\widehat{LSE}}^2  \frac{ \displaystyle \sum_{i = 1}^{K} e^{\alpha d_i } n_i}
{\displaystyle \sum_{i = 1}^{K} e^{\alpha d_i }}
\frac{r}{\displaystyle \| x \|^3}
\big( \| x \|^2 - xx^T- \frac{\| x \|^3}{2r} \big).
\end{equation}
%\begin{equation}
%g_{scp}=
%\frac{\partial \mathcal{V}}{\partial x} = 
%6 \lambda {\widehat{LSE}}^2  \frac{  \sum_{i = 1}^{K} e^{\alpha d_i } n_i}
%{ \sum_{i = 1}^{K} e^{\alpha d_i }}
%\frac{r}{ \| x \|^3}
%\big( \| x \|^2 - xx^T- \frac{\| x \|^3}{2r} \big)
%\end{equation}
We will employ the formation (\ref{star_opt_uncon}) for visibility planning in the following sections.

{\bf Remark.} The formulation (\ref{star_opt_uncon}) is an appropriate adaptation of (\ref{star_opt}) for optimization efficiency and the hard constraint in (\ref{star_opt}) guarantees the visibility. Although (\ref{star_opt_uncon}) shares a similar cost structure with other visibility planners \cite{bonatti2018autonomous, wang2021visibility, jeon2020integrated }, they are essentially different. The trajectory optimization formulation in \hyperref[traj_opt_sec]{Sec. \ref{traj_opt_sec}} allows we take extremely large value for the visibility penalty weight while  \cite{bonatti2018autonomous, wang2021visibility, jeon2020integrated } can not. Otherwise, they will result in non-smooth and less efficient trajectories. 
%\begin{equation}
%\frac{\partial \mathcal{V}}{\partial x} = \left\{
%\begin{array}{ll}
%0 & \widehat{LSE} \le 0 \\
%3 \lambda {\widehat{LSE}}^2()  \frac{ \displaystyle \sum_{i = 1}^{K} \exp^{\alpha(n_i^T(p - a_i))} n_i^T}{de}
%
% & \widehat{LSE} > 0 
%\end{array}
%\right.
%\end{equation}

%\begin{equation}
%\begin{aligned}
%\widehat{LSE} \big(x | \mathbb{S}_i \big) = \frac{1}{\alpha}log \big(e^{\alpha (n_i^T(F(x) - a_i))}+ \\ 
%\cdots + e^{\alpha (n_i^T(F(x) - a_i))} \big) - d_{min}.
%\end{aligned}
%\label{lsehat}
%\end{equation}

%\begin{equation}
%P(i,j) = \sum_{x=|N(V_i) \cap V_j|}^{\min\{|V_j|, |N(V_i)|} 
%\frac{\displaystyle \binom{|V_j|}{x} \binom{|V - V_j|}{|N(V_i)| - x}}
%{\displaystyle \binom{|V|}{|N(V_i)|}}   
%\end{equation}

\section{Visibility Guaranteed Planner}
\begin{figure}
 	\vspace{0.2cm}
 	\centering
 	\includegraphics[width=0.8\linewidth]{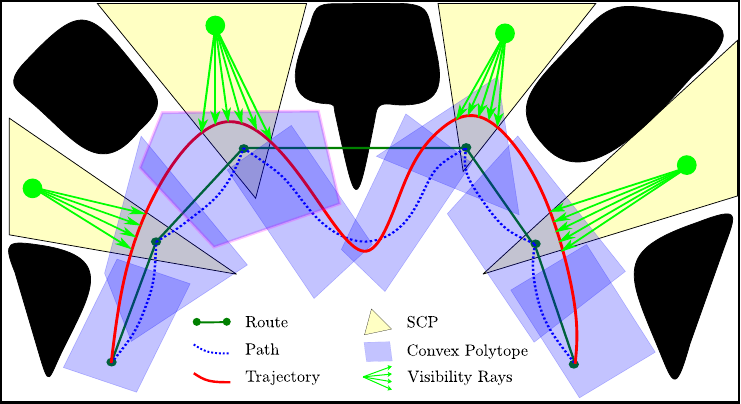}
 	\caption{The whole pipeline is conducted in three steps: 1) route generation and refinement 2) path finding 3) SVC construction 4) trajectory optimization.}
 	\label{principle2}
 \end{figure}

%\vspace{0.8cm}
As shown in the \hyperref[principle2]{Figure \ref{principle2}}, a complete pipeline of visibility guaranteed planner is presented in this section. 
Due to the differential flatness property of multicopters, we can optimize the trajectory in the space of the selected flat output (i.e. the translation of the center of mass and the Euler-yaw angle). To facilitate the trajectory optimization, the map $ \mathcal{M}_g $ is modified by one-to-eight cubic inflation with regards to each point in it. Thus, the map encode the configuration space, and the SCP generated on it can be directly employed as the visibility constraint for trajectory optimization.

% To facilitate the trajectory optimization, the map $ \mathcal{M}_c $ is constructed by the one-to-eight cubic inflation of the points in the map of $ \mathcal{M}_g $. 

%without the euclidean-to-configuration transformation $ \mathcal{P} $.

\subsection{Route Generation and Refinement}
In order not to introduce binary variables to the whole problem, a reasonable visiting sequence of the spots can be obtained in advance by solving the traveling salesman problem (TSP).  Similar to \cite{zhou2021fuel}, we model it as a standard Asymmetric TSP (ATSP) that can be solved efficiently by Lin–Kernighan heuristic (LKH) \cite{helsgaun2000effective}.
%The visiting sequence is indeed an open-loop chain, which is different from the standard TSP formulation. Similar to \cite{zhou2021fuel}, we model it as a standard Asymmetric TSP (ATSP) that can be solved efficiently by Lin–Kernighan heuristic(LKH) \cite{helsgaun2000effective}.
%With the start and end postion $ p_s, p_f $ and visiting spots list $ \mathbb{C} $, we build the cost matrix $ M_{tsp} \in \mathbb{R}^{N+1} $ as
%%\begin{equation}
%%
%%\right.
%%\label{m_tsp}
%%\end{equation}
%\begin{equation}
%M_{tsp}(i, j)=\left\{
%\begin{array}{ll}
% 0                      & i = 0, \, j = 0\\
% \| c_i - c_j\|        & i, \, j \in \{1, \cdots, N \}  \\
% \| p_s - c_j \|        & i = 0, \,  j \in \{1, \cdots, N\} \\ 
% \| p_f - c_i \|        & i \in \{1, \cdots, N\}, \, j=0 
%\end{array} \right. 
%\end{equation}
%where the spot $ p_f $ is regarded as identical to $ p_s $ and the costs from $ \mathbb{C} $ to $ p_s $ are calculated by the euclidean distance between $ \mathbb{C} $ and $ p_f $. Then, the visiting sequence of the spots can be obtained by solving the TSP problem with
%$ M_{tsp} $.
%Since trajectory is not necessary to be closest to the visiting spots as long as it is in the visible space, 
we further optimize route waypoints $ \{w_i  \in \mathbb{R}^3 | i = 1, \cdots, N \} $ on the SCPs to direct the robot for more efficient trajectory. The problem is formulated as finding the minimum of the sum of length on SCPs:
\begin{equation}
\begin{aligned}
\min_{w_1, \cdots, w_N} \quad & \|p_s - w_1 \| + \| w_N - p_f \| + \sum_{i = 2}^{N} \|w_i - w_{i - 1} \|,    \\ 
\mathrm{s.t.} \quad &  w_i \in \mathcal{S}_i, \, \forall i = 1, \cdots, N.
\end{aligned}
\label{min_len}
\end{equation}
For simplification, $ w_0, w_{N+1} $ are alternatively used for $ p_s $ and $ p_f $ hereafter.
According to (\ref{star_opt_uncon}), we further make a relaxation of (\ref{min_len}) to convert it to an unconstrained NLP (nonlinear programming) with cost function
\begin{equation}
\begin{gathered}
J_w =  \sum_{i = 1}^{N} \sqrt{ \| w_i - w_{i - 1} \|^2 + \epsilon } 
 + \Lambda^T \sum_{i = 1}^{N} \mathcal{V}(\widehat{LSE}\big(w_i | \mathcal{S}_i \big) ), 
\end{gathered}
\end{equation}
where the $ \Lambda = [\lambda_1, \cdots, \lambda_N]^T \in \mathbb{R}^N $ is the penalty weight vector and $ \epsilon $ is a small value number for $ C^2 $ condition. By utilizing the previously-derived gradient $ g_{scp} $, the gradient propagation of $ J_w $ can be obtained for $ w_1, \cdots, w_N $:
\begin{equation}
\frac{\partial J_w}{w_i} = \frac{\| w_i - w_{i - 1} \|}
{\displaystyle \sqrt{\| w_i - w_{i - 1} \|^2 + \epsilon } } - 
\frac{\| w_{i + 1} - w_{i} \|}
{\displaystyle \sqrt{\| w_{i + 1} - w_{i} \|^2 + \epsilon } } + 
\lambda_i g_{scp}.
\end{equation} 
Then, the route  $ w_0 \rightarrow w_1 \rightarrow w_2 \rightarrow \cdots, \rightarrow w_N \rightarrow w_{N+1} $ can be obtained by combining the problem of TSP with the optimization problem (\ref{min_len}), which makes preparation for the corridor construction afterwards.
%\label{min_len_unconstrain}

%the trajectory is not necessay to be closest to the visiting spots as long as it is in the visible space

\subsection{Safe and Visible Corridor Construction}
The route generated is not collision free but provide promising flight directions. The route waypoints serve as local goals for kinodynamic A* to search for a collision free path.  We convert the point cloud map to voxel map and perform the search on it, which can save orders of time.

%Comparing to the $ O(1) $ time complexity of query on the voxel map, the query on the k-d tree takes $ O(logN) $ time, which makes kinodynamic A* search on the point cloud map time-consuming, espesically when the map scale is large. 
Based on the searched path, the SVC can be constructed incrementally by connecting the SCPs by sequences of overlapping convex polytopes. For the convex polytope generation, we adopt the efficient method presented in \cite{zhong2020generating} which directly makes modifications to SCP. Consequently, the elements of the corridor can be organized in a unified struct. The intersection between the path and convex polytope is calculated by recursively subdividing the B\'{e}zier form of the trajectory and checking the control points of it. 
For the SCP, the intersection between the path and it can be found via the Point-In-SCP test. 
The convex polytope is built at the intersection until it reaches the next waypoint. Note that we add some augment points to separate the $ j^{th} $ and the  $ (j+2)^{th} $ convex polytopes.

\begin{figure*}[h]
	\vspace{0.2cm}
	\begin{center}
		\includegraphics[scale=0.51]{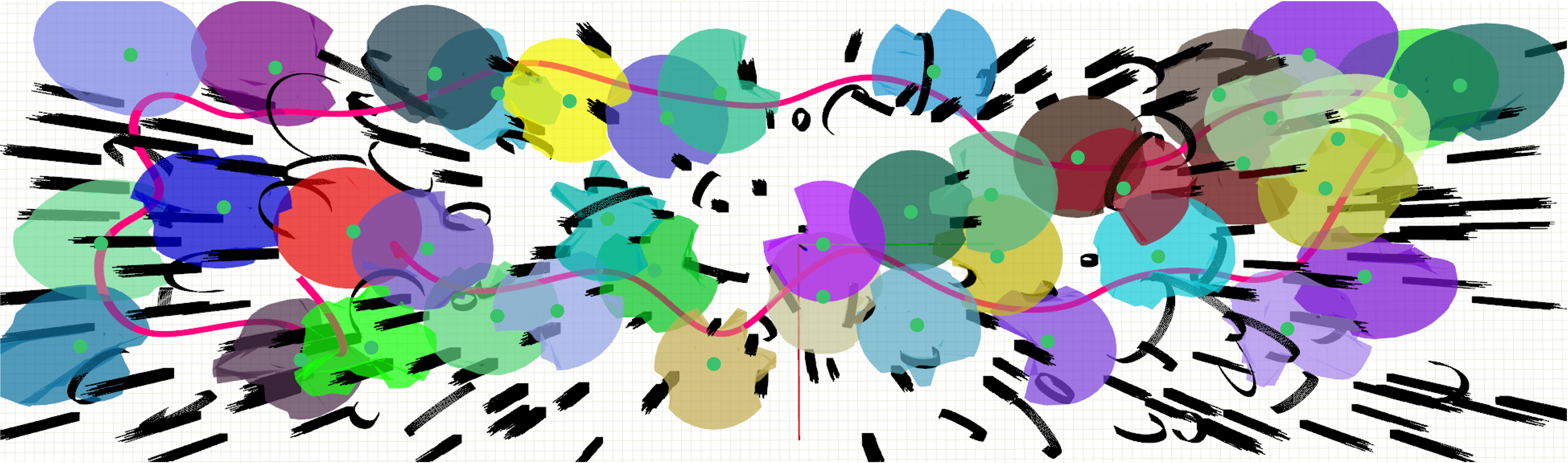}
	\end{center}
	\caption{Simulation in large scene with scale  $ 40 \times 150 \, m $. The scene is composed of 150 pillar-shaped obstacles and 60 ring-shaped ones. There are randomly generated 40 spots for inspection and the corresponding SCPs are shown in different colors. The trajectory is generated in 4.4s and is guaranteed to inspect all the sites. }
	\label{large_scale}	
\end{figure*}

% &\boldsymbol{\sigma^{i}(t) \in S_i, \, \mathcal{T}(\sigma^{i}) > \tau_i} , \nonumber  \\
% &\boldsymbol{\forall i = 1, 2, \cdots, N } , \tag{\ref{trajopt}{d}} \label{trajopt_d}  

\subsection{Trajectory Optimization}
\label{traj_opt_sec}
Given the constructed SVC,
the trajectory generation problem can be formulate as the following time-spatial optimization problem:
 
 \begin{align}
 \min_{\sigma(t)} \; & \int_{0}^{T_{\Sigma}} \| \sigma^{(3)}(t) \|^{2} dt + \rho T_{\Sigma}, \label{trajopt}     \\ 
 \mathrm{s.t.} \quad & [\sigma(0), \sigma^{(1)}(0), \sigma^{(2)}(0) ] = [p_s, v_s, a_s], \tag{\ref{trajopt}{a}} \label{trajopt_a}   \\ 
 & [\sigma(T_{\Sigma}), \sigma^{(1)}(T_{\Sigma}), \sigma^{(2)}(T_{\Sigma}) ] = [p_f, v_f, a_f], \nonumber \\
 & \sigma(t) \in \mathcal{F}, \quad  \forall t \in [0, T_{\Sigma}],  \tag{\ref{trajopt}{b}} \label{trajopt_b} \\
 & \| \sigma^{(1)}(t) \| \le v_m, \| \sigma^{(2)}(t) \| \le a_m, \forall t \in [0, T_{\Sigma}],  \tag{\ref{trajopt}{c}} \label{trajopt_c}  \\
 %& \forall i = 1, 2, \ldots, N, \exists t_a, \, t_b, \,  0 < t_a < t_b < T, \,  \\ 
 %& \quad  t_b - t_a > \tau_i, \, such \: that \: \sigma(t) \in S_i, \forall t\in[t_a, t_b]  \\
 &\boldsymbol{\sigma^{i}(t) \in S_i, \, T_i > \tau_i} , \nonumber  \\
 &\boldsymbol{\forall i = 1, 2, \cdots, N } , \tag{\ref{trajopt}{d}} \label{trajopt_d}  
 \end{align}
where $ \sigma(t): \mathbb{R} \mapsto \mathbb{R}^3 $ is a polynomial spline over $ [0, T_{\Sigma}] $ with time allocation $ [T_1, T_2, \cdots, T_N] $ on SCPs,
$ T_{\Sigma} $ the total time of $ \sigma(t) $, $ \rho $ the time regularization weight. The trajectory is constrained to be collision free, dynamic feasible, and visibility capable, which corresponds to the conditions in (\ref{trajopt_b}), (\ref{trajopt_c}) and (\ref{trajopt_d}) respectively. Then,  we denote by $ \mathcal{F} $ the resultant safe and visible corridor, $ v_m $ and $ a_m $ the dynamic limits, 
$ \sigma^{i}(t) $ the segment of $ \sigma(t) $ that is assigned to the $ i^{th} $ SCP.

To solve the optimization problem (\ref{trajopt}), we generally adopt the directly constructed minimum control trajectory \textit{MINCO} from \cite{wang2021geometrically}. Similar to \cite{wang2021geometrically}, 
smooth maps are utilized to exactly eliminate spatial and time constraints. The dynamic constraint (\ref{trajopt_c}) is transformed into a finite-dimensional one via integral of
constraint violation. For brevity, we refer reader to \cite{wang2021geometrically} for more details.

For the star-convex constraint in  (\ref{trajopt_d}), we make a relaxation via integral of constraint violations. According to (\ref{star_opt_uncon}), we eliminate the constraint by
% (\ref{scp_v}) and (\ref{lsehat}),
 defining the time integral penalty for visibility:
\begin{equation}
I(\mathcal{S}_i, \eta_i) = \frac{T_i}{\eta_i} \sum_{j = 0}^{\eta_i } \mathcal{V} \Big( \widehat{LSE} \big(\sigma(j \frac{T_i}{\eta_i} ) | \mathcal{S}_i \big)
\Big),
\label{vis_intgral_pen}
\end{equation}
where $ T_i $ is the time for the $ i^{th} $ segment of the trajectory, and $ \eta_i $ 	controls the relative resolution of the quadrature. For the minimum time constraint in (\ref{trajopt_d}), we take the decision variable mapping
\begin{equation}
T_i = e^{\xi_i} + \tau_i,
\label{t_diffeo}
\end{equation}
 to eliminate the constraint as well, where $ \xi = (\xi_1, \cdots, \xi_N) $ is $ C^{\infty} $ diffeomorphic to 
$ T = (T_1, \cdots, T_N) $ . By incorporating (\ref{vis_intgral_pen}) and (\ref{t_diffeo}) into the optimization framework \cite{wang2021geometrically}, the optimization problem (\ref{trajopt}) can be transformed into the unconstrained control effort minimization problem which can be solved efficiently and reliably. %The searched trajectory serves as the initial value for the optimization.

% searched acceleraton controlled trajectory serves as the initial value for the NLP problem.
\section{Application On Aerial Inspection }
Our planner can be employed for exploration, tracking, and many other applications with task-specific modification.  To best evaluate our planner and  motivated by the need to regularly inspect factories or substations, we test our planner under site inspection background, where our planner can be employed without extra effort. The task requires that a drone can observe every spot for enough time while saving the task time and energy as much as possible.
%\begin{figure}
%	\centering
%	\includegraphics[scale=0.23]{figures/sim_env.png}
%	\caption{The random environment used for simulation.}
%	\label{sim_env}
%\end{figure}
% as is shown in \hyperref[sim_env]{Figure \ref{sim_env}}
\subsection{Simulation and Benchmark Comparisons}
We test the proposed method in a randomly generated environment consisting of pillar-shaped and ring-shaped obstacles. 
%To demonstrate the superiority of our method in various environments with different scales,
To demonstrate the scalability of the method, three scenarios are designed with increasing problem scale:
\begin{itemize}
	\item \textit{Small}: $ 20 \times 20\, m $, $ 15 $ pillars and $ 6 $ rings, $ 3 $ spots.
	\item \textit{Medium}:  $ 40 \times 40 \, m $, $ 60 $ pillars and $ 20 $ rings, $ 10 $ spots.
	\item \textit{Large}: $ 80 \times 80 \, m $, $ 150 $ pillars and $  60 $ rings, $ 20 $ spots.
\end{itemize}
We set the dynamic limits of drone as $ v_{max} = 4.0 \, m/s $ and $ a_{max} = 6.0 \, m/s $. 
All the simulations are conducted with a 2.6 GHz Intel i7-9750H processor.

\begin{figure}[h]
	\centering
	\includegraphics[scale=0.35]{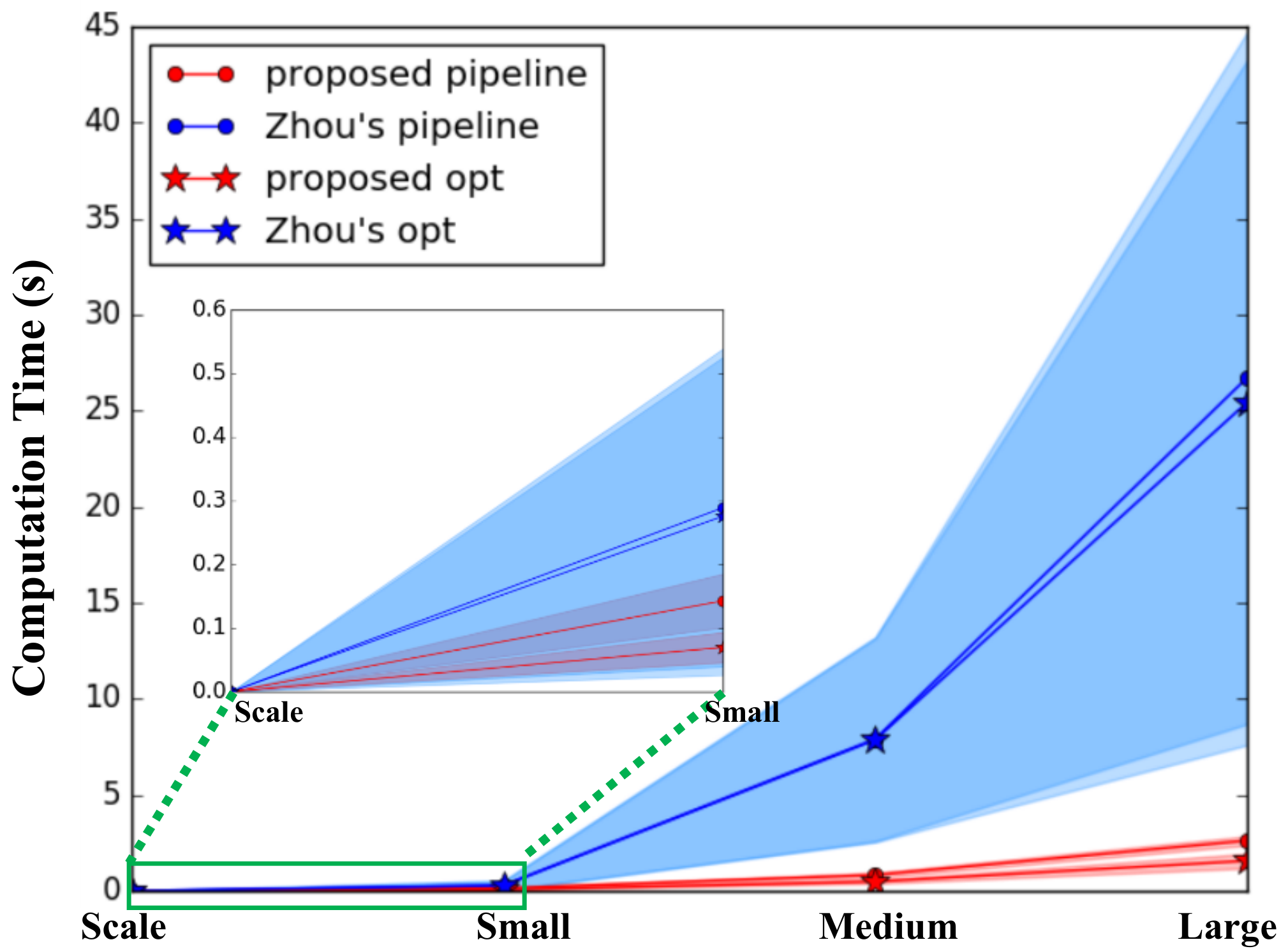}
	%	\hspace{1in}
	\caption{Benchmark comparison of the computational time for different scales (The ESDF construction time is not counted in Zhou's). Both the trajectory optimization and the pipeline time are evaluated. The shaded area is the $ 4/5 \sigma $ interval, where $ \sigma $ is the standard deviation.}
	\label{cp_time}
\end{figure}

In the implementation, we set $ R = 6.0 \, m $ to confine the SCP in a ball, $ r = 20 \, m $ for ball flipping. In the trajectory generation, we use, $ \rho = 150 $, $ \eta_i = 10 $. For the LSE function, we set $ \alpha = 100.0 $, which can make an approximation with the precision of 0.01. We benchmark the method with \cite{zhou2021fuel} whose planning framework is similar to ours (i.e. TSP + trajectory optimization). Although \cite{zhou2021fuel} is a local planner, our evaluation spectrum covers  both the local and global scales. The computation time (\hyperref[cp_time]{Figure \ref{cp_time}}) of our method shows it is also adequate for replanning. For Zhou's method, we make a few modifications to fit into our application. Firstly, the frontier information structure is left out because TRPs are already given in site inspection. The TRPs are equivalent to the average positions of frontier clusters in Zhou's method. Secondly, the viewpoints are generated without considering  the yaw angle and occlusion effect is degraded to  points connectivity to comply with the omnidirectional sensor assumption.
%the space of every spot is discretized by $ \{0.5 \, rad, 0,5 \, rad, 0.5 \, m\}  $ in spherical coordinate system and the visible points are checked by raycasting for the omnidirectional sensor assumption. 
Thirdly, the route is generated and refined by constructing a graph on visible points by Euclidean distance instead of the path length searched by A*, for the reason of saving computation time.

% For the b-spline configuration, we set the segemnt length as $ 6.0m $ and control points distance as $ 1.0m $.

\begin{table}
	\centering
	\fontsize{8}{11} \selectfont
	\caption{Statistic on trajectory quality} 
	\begin{tabular}{ccccc}
		\toprule
		\toprule
		Scene Scale & Method & Traj dur (s) & Int ($ J^2 $) & Vis cap \cr
		\multirow{2}{*}{Small} & Zhou et. al & \textbf{6.9}  & 546.5 & 80\%  \cr
		& Proposed & 7.7  & \textbf{159.3} & \textbf{100\%}    \cr
		\midrule
		\multirow{2}{*}{Medium} & Zhou et. al & \textbf{31.4} & 819.5 & 54\%  \cr
		& Proposed & 32.8 & \textbf{371.2} & \textbf{100\%}   \cr	
		\midrule
		\multirow{2}{*}{Large} & Zhou et. al & \textbf{63.5} & 1063.9 & 37\%   \cr
		& Proposed & 65.3 & \textbf{487.8} &  \textbf{100\%}   \cr	
		\bottomrule
		\bottomrule		   
	\end{tabular}
	\label{bench_res}
\end{table}	

\begin{figure}
	\vspace{0.2cm}
	\centering
	\includegraphics[scale=0.4]{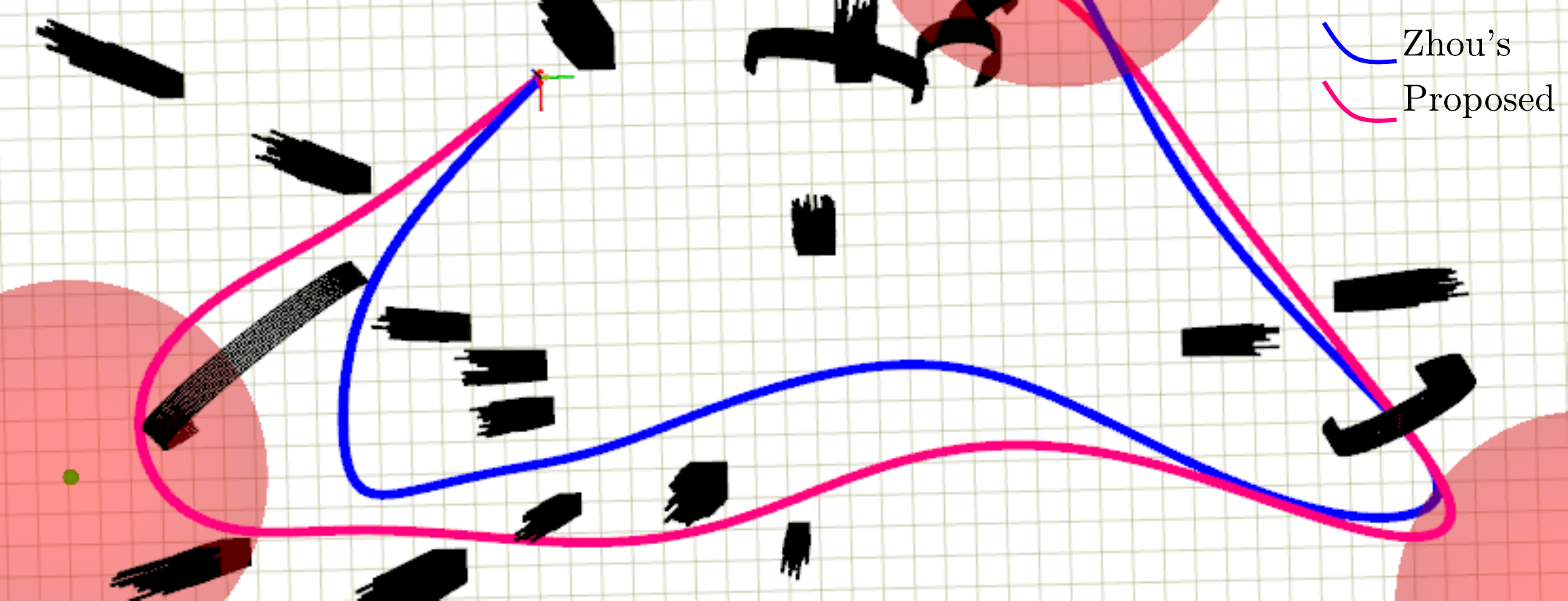}
	%	\hspace{1in}
	\caption{Comparison of the generated trajectory. The red ball is the sensible region for each spot. The trajectory planned by Zhou's method failed to inspect the left bottom spot. }
	\label{comp_traj}
\end{figure}
The \hyperref[bench_res]{Table \ref{bench_res}} shows the statistics on the trajectory quality. The visible capability refers to the ratio of observed spots. Owing to the SCP and the corresponding constraint formulation, our method can guarantee the visibility of all the spots, while Zhou's method loses many hits for them. In addition, our method is more smooth and energy-efficient, indicated by the criterion of Int($ J^2 $) (time integral of squared jerk). This primarily benefits from the powerful trajectory optimization framework \cite{wang2021geometrically}. The optimized trajectory duration is higher than Zhou's but still comparable to it.  Without the hard visibility constraint, Zhou's method tends to reduce the length of trajectory, which will reduce the execution time, as shown in \hyperref[comp_traj]{Figure \ref{comp_traj}}.

%\begin{figure}
%	\centering
%	\includegraphics[scale=1.0]{figures/cp_slim.pdf}
%	%	\hspace{1in}
%	\caption{Timing breakdown for the proposed pipeline in the medium scene.}
%	\label{box_all}
%\end{figure}
The comparison of the computational time is shown in \hyperref[cp_time]{Figure \ref{cp_time}}. Our method is faster than Zhou's by orders of magnitudes and is more reliable.
Lacking a compact environment abstraction (e.g. SVC), the trajectory optimization time of Zhou's takes almost $ 99\% $ of the whole pipeline. The proposed method spends about $ 42\% $ of total time for the generation of SCPs, route, and SVC, but they highly speed up trajectory optimization. As the problem scale increase, our method can still finish in seconds. A more large scale test of the proposed method is shown in \hyperref[large_scale]{Figure \ref{large_scale}}.

%\begin{table}
%	\centering
%	\fontsize{8}{11} \selectfont
%	\caption{This is a visual sample.} 
%	\begin{tabular}{ccccccccc}
%		\toprule
%		\toprule
%		\multirow{2}{*}{Scale} & \multirow{2}{*}{Method} & \multicolumn{3}{c}{Traj dura (s)} & \multicolumn{3}{c}{Int ($ J^2 $)} & \multirow{2}{*}{Vis cap}  \cr
%		\cmidrule{3-8}
%		& &  Avg & Max & Min & Avg & Max & Min \cr	
%		\multirow{2}{*}{Small} & Zou et. al & 6.9 & 26.6 & 4.3 & 546.5 & 884.3 & 240.3 & 80\% \cr
%		& Proposed & 7.7 & 14.6 & 4.3 & 159.3 & 269.7 & 75.7 & 100\%  \cr
%		\midrule
%		\multirow{2}{*}{Medium} & Zou et. al & 1 & 2 & 3 & 1 & 2 & 3 & \cr
%							& Proposed & 1 & 2 & 3 & 1 & 2 & 3 & 100\%  \cr	
%		\midrule
%		\multirow{2}{*}{Large} & Zou et. al & 1 & 2 & 3 & 1 & 2 & 3 & \cr
%								& Proposed & 1 & 2 & 3 & 1 & 2 & 3 & 100\% \cr	
%		\bottomrule
%		\bottomrule		   
%	\end{tabular}
%	\label{bench_res}
%\end{table}

%[tbp]

%\begin{figure}
%	\centering
%	\includegraphics[scale=0.3]{figures/scp-con.png}
%	%	\hspace{1in}
%	\caption{Illustaration of star convex polytope construction in 2D. Point transformation: the points(green) inside the sphere is transfered to the points outside of the sphere(blue) by one to one correspondence. Convex hull construction: the convex hull is constructed in the nonlinear space outside of the sphere; the blue dotted curve is the image of it in the original space.   }
%	\label{scp}
%\end{figure}

\subsection{Real-World Experiment}

\begin{figure}[h]
	\vspace{0.2cm}
	\centering
	\includegraphics[width=0.9\linewidth]{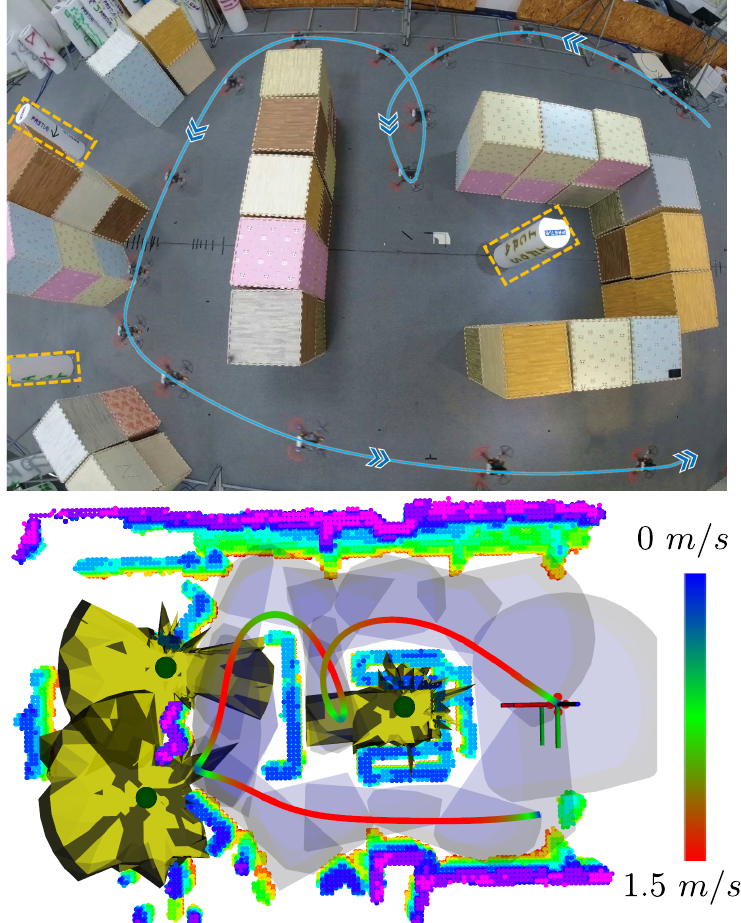}
	\caption{Real world scene to test the proposed method. Refer to \hyperref[front_test]{Figure \ref{front_test}} for the labels. The color map indicates the velocity of the quadrotor.}
	\label{exp}
\end{figure}
%\begin{figure}[h]
%	\centering
%	\includegraphics[width=1.0\linewidth]{figures/color_traj.png}
%	\caption{Real world scene to test the proposed method.}
%	\label{exp_real}
%\end{figure}

We conduct real-world indoor experiment to validate the proposed method, as shown in \hyperref[exp]{Figure \ref{exp}}. The upright cylindrical obstacles are the targets to be inspected. The map is pre-built  using lidar by LIO-SAM \cite{legoloam2018shan} and the trajectory is planned offline. 
The quadrotor we used is equipped with an Intel Realsense D435 for state estimation and Insta 360 One X2\footnote[1]{https://www.insta360.com/} for omnidirectional perception. The maximum velocity and acceleration are set as $ 1.5\, m/s $ and $ 1.0 \, m/s^2 $. The minimum inspection time for each object is set as $ 1.0 s$. 

The test environment and the associate results are displayed in \hyperref[exp]{Figure \ref{exp}}. The quadrotor is able to inspect all the targets. Since 
the quadrotor is not necessary to be closest to the targets as long as they are visible, it slows down and inspect the target through the gap. The test shows that  the SCP can excavates almost all visible regions and the formulated \textit{star-convex constrained optimization} renders reasonable trajectory for visibility planning.
%The quadrotor platform is  equipped with a
%Velodyne VLP-16 3-D LiDAR. LOAM [20] is adopted to
%estimate the pose of the quadrotor and generate a dense point
%cloud map

\section{Conclusion}
In this paper, we introduce
 a compact and efficient space representation the SCP and propose to formulate the visibility constraint for \textit{star-convex constrained optimization}. By utilizing the SCP, we design a visibility guaranteed planning framework, while retains the safety, feasibility, and energy efficiency of trajectory. The experimental results show that the method is efficient, scalable, and visibility guaranteed.
 
 The main limitation of our method the is the omnidirectional perception assumption of the sensor model. In the future, we will take limited FOV of sensors into consideration and plan the yaw angle in trajectory optimization.
 
% \section*{ACKNOWLEDGMENT} 
% This work was supported by the National Natural Science Foundation of China under Grants 62003299. The authors would also like to thank Botao He for his help with the hardware experiment.
 
%\begin{table}
%	\centering
%	\caption{Run time under different obstacle density}
%	\begin{tabular}{cccc}
%		\toprule
%		\toprule
%		Obstacle number & 1 & 40 & 60 \\
%	    Average of Run time(s) & 0.23 & 0.52 & 1.08\\
%
%		\bottomrule
%		\bottomrule
%	\end{tabular}\vspace{0cm}
%	\label{tab:Training_sizes}
%\end{table}

%\begin{figure}[h]
%	\centering
%	\includegraphics[width=1\linewidth]{figures/sfc.png}
%	\caption{Left: the line segments searched by jump point search and the initialized ellipsoid. Right: the union of polyhedrons constructed by manipulating the ellipsoid to model the safe flight corridor.}
%	\label{sfc}
%\end{figure}

%\begin{algorithm}[h]
%	\SetAlgoLined
%	\KwResult{Write here the result }
%	initialization\;
%	\While{While condition}{
%		instructions\;
%		\eIf{condition}{
%			instructions1\;
%			instructions2\;
%		}{
%		instructions3\;
%	}
%}
%\caption{How to write algorithms}
%\end{algorithm}

\addtolength{\textheight}{0cm}   % This command serves to balance the column lengths

%%%%%%%%%%%%%%%%%%%%%%%%%%%%%%%%%%%%%%%%%%%%%%%%%%%%%%%%%%%%%%%%%%%%%%%%%%%%%%%%

%%%%%%%%%%%%%%%%%%%%%%%%%%%%%%%%%%%%%%%%%%%%%%%%%%%%%%%%%%%%%%%%%%%%%%%%%%%%%%%%

%%%%%%%%%%%%%%%%%%%%%%%%%%%%%%%%%%%%%%%%%%%%%%%%%%%%%%%%%%%%%%%%%%%%%%%%%%%%%%%%
%\section*{APPENDIX}

%\section*{ACKNOWLEDGMENT}

%The preferred spelling of the word `acknowledgment' in America is without an `e' after the `g'. Avoid the stilted expression, `One of us (R. B. G.) thanks . . .'  Instead, try `R. B. G. thanks'. Put sponsor acknowledgments in the unnumbered footnote on the first page.

%%%%%%%%%%%%%%%%%%%%%%%%%%%%%%%%%%%%%%%%%%%%%%%%%%%%%%%%%%%%%%%%%%%%%%%%%%%%%%%%

%References are important to the reader; therefore, each citation must be complete and correct. If at all possible, references should be commonly available publications.
%They are cited like so: \cite{IEEEexample:articleetal}, \cite{IEEEexample:book}, ...
%\newpage

\bibliographystyle{IEEEtran}
\bibliography{IEEEabrv,test}

\end{document}